\pdfoutput=1

\documentclass[11pt]{article}

\usepackage[]{EMNLP2023}

\usepackage{times}
\usepackage{latexsym}

\usepackage[T1]{fontenc}

\usepackage[utf8]{inputenc}
\usepackage{arabtex}
\usepackage{utf8}

\usepackage{microtype}
\usepackage{inconsolata}
\usepackage{multirow}
\usepackage[spacesep,definevectors]{easyvector}
\usepackage{amsmath}
\usepackage{mathtools}
\usepackage{xspace}
\usepackage{algpseudocode}
\usepackage{algorithm}
\usepackage{hyperref}

\DeclareMathOperator*{\argmin}{arg\,min}
 
\newcommand{\OurMODEL}{GARI} 

\newcommand{\eat}[1]{}

\renewenvironment{abstract}%
         {\centerline{\large\bf Abstract}%
          \begin{list}{}%
             {\setlength{\rightmargin}{0.6cm}%
              \setlength{\leftmargin}{0.6cm}}%
           \item[]\ignorespaces}%
         {\unskip\end{list}}

\title{GARI: Graph Attention for Relative Isomorphism of Arabic Word Embeddings}

\author{
    Muhammad Asif Ali,\textsuperscript{\rm 1}
    Maha Alshmrani,\textsuperscript{\rm 1}
    Jianbin Qin,\textsuperscript{\rm 2}
    Yan Hu,\textsuperscript{\rm 1}
    Di Wang\textsuperscript{\rm 1}\\
    \textsuperscript{\rm 1} King Abdullah University of Science and Technology, KSA\\
    \textsuperscript{\rm 2} Shenzhen University, China\\
    \{muhammadasif.ali; maha.shmrani; yan.hu; di.wang\}@kaust.edu.sa; qinjianbin@szu.edu.cn\\
}

\begin{document}

\maketitle
\begin{abstract}
Bilingual Lexical Induction (BLI) is a core challenge in NLP, it relies on the relative 
isomorphism of individual embedding spaces. Existing attempts aimed at controlling the 
relative isomorphism of different embedding spaces fail to incorporate the impact of 
semantically related words in the model training objective. To address this, we propose 
\OurMODEL~that combines the distributional training objectives with multiple isomorphism 
losses guided by the graph attention network. \OurMODEL~considers the impact of semantical 
variations of words in order to define the relative isomorphism of the embedding spaces. 
Experimental evaluation using the Arabic language data set shows that \OurMODEL~outperforms 
the existing research by improving the average P@1 by a relative score of up to 40.95\% 
and 76.80\% for in-domain and domain mismatch settings respectively. We release the codes 
for \OurMODEL~at \url{https://github.com/asif6827/GARI}.
\end{abstract}

\section{Introduction}
\setcode{utf8}
\label{intro}

Bilingual Lexical Induction (BLI) is a key task in natural language processing.
It aims at the automated construction of translation dictionaries 
from monolingual embedding spaces.
BLI plays a significant role in multiple different natural language processing applications.
For instance, the automated construction of lexical dictionaries plays a key role in the 
development of linguistic applications for low-resource languages, especially in 
cases where hand-crafted dictionaries are non-existent.
Automated construction of high-quality dictionaries also helps in augmenting the 
end performance of multiple down-streaming tasks, including but not limited to: 
machine translation~\citep{2018_lample}, information retrieval~\citep{2018_artetxe},
cross-lingual transfers~\citep{2019_artetxe_massive}.

Earlier methods aimed at the construction of cross-lingual embeddings use 
linear and/or non-linear mapping functions in order to map the monolingual 
embeddings in a shared space. Some examples in this regard include retrieval criteria 
for bilingual mapping by \citet{2018_joulin} and BLI in non-isomorphic spaces by~\citet{2019_patra}.

These methods rely on the approximate isomorphism assumption, i.e., they 
assume that underlying monolingual embedding spaces are geometrically similar, 
which severely limits their use to closely related data sets originating from 
similar domains and/or languages exhibiting similar characteristics.
The limitations of the mapping-based methods, especially their inability to handle 
data sets originating from different domains and languages exhibiting different 
characteristics has been identified by~\cite{muse_2017,2018_sogaard,2019_glavas,2019_patra}.

Some other noteworthy aspects identified in the literature that limit the 
end performance of the BLI systems, include: 
(a) algorithmic mismatch for independently trained monolingual embeddings,
(b) different parameterization,
(c) variable data sizes, 
(d) linguistic difference, etc.,~\citep{2020_marie_iter,2022_isovec}.

\begin{figure}[t]
	\centering
	\includegraphics[width=0.76\linewidth]{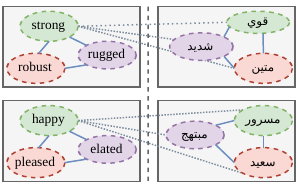}
	\caption{Some examples of semantically related tokens for English and their corresponding translations in 
		the Arabic language.}
	\label{fig:iso_example}
	\vspace{-3.7ex}
\end{figure}

In the recent past, there has been a shift in the training paradigm for the BLI 
models, i.e., instead of relying on pre-trained embeddings trained independently of each other, they use explicit isomorphism metrics along 
with the distributional training objective~\citep{2022_isovec}.
However, a key limitation of these models is their inability to 
incorporate the impact of semantically related tokens (including their lexical 
variations) in controlling the relative isomorphism of different spaces.
This is illustrated in Figure~\ref{fig:iso_example}, where the left half 
of the figure shows a set of semantically related English words, e.g., \{strong, rugged, and robust\}.
These words though lexically different share the same semantics. Correspondingly, their translations in 
the Arabic language: \{\<شديد, قوي, متين>\} are also semantically related.
We hypothesize that each language encompasses a list of such semantically related 
words that may be used interchangeably within a fixed context, and in order to control 
the relative isomorphism of corresponding embedding spaces the end model should be robust to 
incorporate these semantic variations in the model training objective.

\eat{This work emphasizes that in order to control the relative isomorphism of the embedding spaces there 
	is a need to constraint the embedding vectors for the translation pairs in a way that embedding vectors
	for all semantically related words end up being close to each other.}

To address these challenges, in this paper, we propose Graph Attention for Relative Isomorphism 
(\OurMODEL). \OurMODEL~combines the distributional training objective with the isomorphism loss
in a way that it incorporates the impact of semantically related words using graph attention, 
required to perform the end-task in a performance-enhanced way. 
We outline the key contributions of this work as follows:

\begin{enumerate}
	\item We propose~\OurMODEL~that combines the distributional loss with graph attention-based 
	isomorphism loss functions for effective BLI.
	\item The graph attention part of the \OurMODEL~leverages self-attention mechanism in order to
	attend over words that are semantically related to a given word.
	\item We prove the effectiveness of \OurMODEL~by comprehensive experimentation. Experimental 
	evaluation shows, for the Arabic dataset, the \OurMODEL~outperforms the existing research on 
	relative isomorphism by 40.95\% and 76.80\% for in-domain and out-of-domain settings.
\end{enumerate}

\vspace{-1ex}
\section{Related Work}
\vspace{-1ex}
There is an immense literature on BLI and controlling the relative isomorphism 
of the embedding spaces. In order to save space, we primarily limit the related 
work of this paper to one that is more relevant to our problem settings. 
We classify the related work into the following categories: 
(i) mapping pre-trained embeddings, (ii) combined training.

\paragraph{Mapping Pre-trained Embeddings.}
These methods rely on the use of linear and/or non-linear mappings to map the 
mono-lingual embeddings to a shared space. 

Earlier works in this regard include principled bilingual dictionaries 
by~\citet{2016_artetxe} that aim to learn bilingual mappings while preserving 
invariance for the monolingual analogy tasks.
\citet{2017_artetxe} introduced a self-learning approach to relax the 
requirements for bilingual training seeds and/or parallel corpora.
\citet{2018_alvarez} formulate the alignment as an optimal transport problem
and employ Gromov-Wasserstein distance to compute the similarity of word pairs 
across different languages.
\citet{2018_doval} propose additional transformation on top of the alignment 
step to force the synonyms towards a middle point for a better 
cross-lingual integration of the vector spaces.
\citet{2019_Pratik} introduced language-specific rotations followed 
by a language-independent similarity in a common space.
Similar to the word embedding methods, the application of the 
mapping-based methods to the contextualized embeddings include 
context-aware mapping by 
\citet{2019_hanan} and alignment of contextualized embeddings by \citet{2019_schuster}.

\paragraph{Combined Training.}
On contrary to the mapping-based methods that rely on pre-trained embeddings, 
these methods use parallel data as input in order to jointly minimize the mono-lingual as 
well as cross-lingual training objectives.
\citet{2017_duong} introduced methods for cross-lingual word embeddings for multiple
languages in a unified vector space aimed to combine the strengths of different languages.
\citet{2019_wang} addressed the limitations of joint training methods by combining them with 
mapping-based schemes for model training.
For more details on the joint training methods refer to the survey paper by~\citet{2019_ruder}.
\citet{2022_isovec} introduced IsoVec which uses multiple different
isomorphism metrics with skip-gram as the distributional training objective to 
control the isomorphism.

Nevertheless, we observe that existing methods for controlling the relative 
isomorphism ignore the impact of words that are semantically related to a given word,
severely limits the ability of these methods to control the relative isomorphism of 
the embedding spaces.

\vspace{-1ex}
\section{Background}
\vspace{-1ex}
In this section, we first introduce the mathematical notation being used 
throughout the paper and formulate our problem definition. 
Later, we provide a quick background of the VecMap~\citep{2018_artetxe}, 
a toolkit for mapping across different embedding spaces. 

\subsection{Notation}
For this work, we use $\mathbf{X} \in \mathbf{R}^{m \times d}$ and 
$\mathbf{Y} \in \mathbf{R}^{n \times d}$ to represent the embedding 
matrices for the source and target languages with vocab size $m$ and $n$
respectively. $d$ refers to the dimensionality of the embedding space.
The embedding vectors for words, e.g., \{$x, y$\} are represented by
\{$\vec{\mathbf{x}}, \vec{\mathbf{y}}$\}.
Like existing supervised works on controlling the relative isomorphism, e.g., 
IsoVec by \citet{2022_isovec}, we assume the availability of training seeds 
pairs for the source and target languages, denoted by: $\{(x_0,y_0), (x_1,y_1),...(x_s,y_s)\}$.

\subsection{The problem}
In this work, we address a core challenge in BLI, i.e., we control the relative isomorphism of the
embedding spaces. Specifically, we learn the distributional embeddings for the source language (i.e., Arabic) in a way: 
\begin{enumerate}
	\item The source embeddings $\mathbf{X}$ are geometrically isomorphic to the target embeddings $\mathbf{Y}$ 
	(i.e., English language).
	\item While learning isomorphic embeddings the $\mathbf{X}$ should incorporate the impact of the 
	semantically related tokens (also their lexical variations) in $\mathbf{Y}$ in order to perform the end 
	task in a performance-enhanced way. 
\end{enumerate}


\subsection{VecMap toolkit}
\label{vecmap_bkgd}

We use VecMap toolkit\footnote{\url{https://github.com/artetxem/vecmap}}
for mapping across different embedding spaces.
For this, we pre-process the embeddings using a process flow outlined by 
~\citet{2019_zhang_girls}. The embeddings are unit-normed, mean-centered 
followed by another round of unit-normalization.
For bi-lingual induction, we follow~\citep{2018_artetxe}, i.e., whitening 
the spaces, and solving Procrustes.
Later, we perform re-weighting, de-whitening, and mapping of translation 
pairs via nearest-neighbor retrieval~\citep{2018_artetxe}.

\vspace{-1ex}
\section{Proposed Approach}
\begin{figure}[t]
	\centering
	\includegraphics[width=1.01\linewidth]{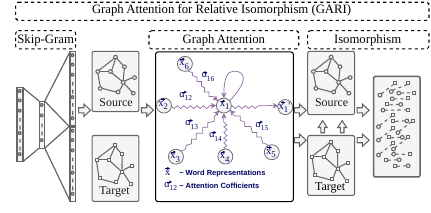}
	\caption{Graph Attention for Relative Isomorphism (GARI), the framework proposed in this work.
		It combines skip-gram and isomorphism loss (guided by graph attention).}
	\label{fig:proposed}
	\vspace{-3.7ex}
\end{figure}

\vspace{-1ex}
In this paper, we address a core challenge in controlling the 
geometric isomorphism for source word embeddings relative to 
the target word embeddings, i.e., incorporate the impact of 
semantically coherent words in order to perform the end task in a 
performance augmented fashion.
For this, we propose Graph Attention for Relative Isomorphism 
(\OurMODEL), shown in Figure~\ref{fig:proposed}. Details about the individual 
components of \OurMODEL~are provided in the following subsections.

\subsection{\OurMODEL}
\subsubsection{Overview}
\OurMODEL~aims to learn the source distributional embeddings $\mathbf{X}$ in a way that: 
(a) $\mathbf{X}$ is geometrically isomorphic to the target embeddings $\mathbf{Y}$,
(b) $\mathbf{X}$ incorporates the impact of semantic variations of words in $\mathbf{Y}$.
In order to control the geometric isomorphism of the embedding spaces in a 
robust way, \OurMODEL~uses graph attention mechanism (to incorporate 
the impact of semantically related tokens) prior to using the isomorphism loss functions.
Finally, it combines the distributional training objective and the
isomorphism loss as the training objectives of the complete model.

\subsubsection{Distributional Representation Learning}
In order to learn the distributional embeddings for \OurMODEL, we 
use skip-gram with negative sampling~\cite{2013_mikolov_sg}. 
Its formulation is shown in Equation~\ref{SG_eq}, i.e, embed a word 
close to its neighboring words within a fixed contextual window, while at 
the same time pushing it away from a list of random words selected 
from a noisy distribution.
\begin{equation}
\label{SG_eq}
\begin{aligned}
\mathcal{L}_{Dis} = \log \sigma({\vec{\mathbf{x}^{'}}_{c_O}}^{\mathsf{T}} \vec{\mathbf{x}}_{c_{I}}) + \\
\sum_{i=1}^{k} \mathbf{E}_{c_i \sim P_{n}(c)} \big[ \log \sigma (-{\vec{\mathbf{x}^{'}}_{c_i}}^{\mathsf{T}} \vec{\mathbf{x}}_{c_{I}})\big] 
\end{aligned}
\end{equation}

Here $\vec{\mathbf{x}}_{c_{O}}$ and $\vec{\mathbf{x}}_{c_{I}}$ correspond to the output and input 
vector representations of the word $c$. $k$ is the number of noisy 
samples and $\vec{\mathbf{x}}^{'}_{c_{i}}$ is the embedding vector for the noisy word
selected from the noisy distribution $P_{n}(c)$.

\subsubsection{Semantic Relatedness}
To incorporate the impact of semantically related words in controlling 
the relative isomorphism of the embedding spaces, \OurMODEL~uses
graph attention mechanism. The graph attention part of \OurMODEL~works as follows:
(a) create a graph $\mathbf{G}$ such that semantically related words end up being neighbors in the graph, 
(b) use graph attention mechanism for information sharing among neighbors in $\mathbf{G}$. 
The details about individual components are as follows:

\paragraph{(a) Graph Construction.} The end goal of the graph construction step is to unite 
and/or combine the semantically related words helpful in controlling the relative isomorphism. 
Inputs for the graph construction process include: 
(i) pre-trained word2vec embeddings\footnote{\url{https://code.google.com/archive/p/word2vec/}, 
	trained using Google-News Corpus of 100 billion words.}, and
(ii) seed words corresponding to the target language, i.e., $\{y_0, y_1, ... ,y_s\}$. 
The graph construction process proceeds as follows:

(a) Organize all seed words for the target language as a set of pairs:
$\mathbf{P} = \{(y_0, y_1), (y_0, y_2), ... ,(y_s,y_s)\}$, i.e., combinations of two words at a time.

(b) For each pair compute the cosine similarity score between the 
corresponding word2vec embedding vectors, and retain only the subset 
($\mathbf{P}_{sub}$) with the cosine similarity score greater than a threshold ($\eta$).

(c) Finally, for the word pairs in $\mathbf{P}_{sub}$ construct a graph $\mathbf{G}$ by 
formulating edges between the word pairs.

Note, this setting for the graph construction allows each word to be surrounded 
by a set of semantically related neighbors which provides \OurMODEL~with the 
provision to allow the propagation of information by using graph attention, as explained below.

\paragraph{(b) Graph Attention.} The graph attention part of \OurMODEL~follows 
a similar approach as proposed by~\citet{2017_velivckovic_GAT}. 
For a graph $\mathbf{G}$, the inputs to a single attention layer of the graph attention network 
include the source word representations 
$\{\vec{\mathbf{x}_{0}},\vec{\mathbf{x}_{1}},...,\vec{\mathbf{x}_{s}} \}, 
\vec{\mathbf{x}_i} \in \mathbf{R}^{d}$, where $s$ represent the number of words 
and $d$ represents the dimensionality of the feature.
It generates a new set of word representations $\{\vec{\mathbf{x}^{'}_{0}},\vec{\mathbf{x}^{'}_{1}},...,\vec{\mathbf{x}^{'}_{s}}\}, \vec{\mathbf{x}^{'}_i} \in \mathbf{R}^{d^{'}}$ as output. Its process flow is explained as follows:

Initially, a linear transformation is applied to all the words in $\mathbf{G}$ 
parameterized by a shared matrix $\mathbf{W} \in \mathbf{R}^{d \times d^{'}}$.
This is followed by using a shared attention mechanism 
$z:\mathbf{R}^{d^{'}} \times  \mathbf{R}^{d^{'}} \rightarrow \mathbf{R}$ 
to compute the intermediate attention coefficients $\beta_{ij}$
that incorporates the importance of word $j$ on word $i$.

\begin{equation}
\beta_{ij} = z(\mathbf{W}\vec{\mathbf{x}_{i}} , \mathbf{W}\vec{\mathbf{x}_{j}})
\end{equation}

where the attention mechanism $z$ is simply a single-layered feed-forward
neural network with a weight vector $\vec{\mathbf{z}} \in \mathbf{R}^{d^{'}}$
and $\text{ReLU}$ non-linearity, as shown below:

\begin{equation}
z = \text{ReLU} \Big( \vec{\mathbf{z}}^{T} [\mathbf{W}\vec{\mathbf{x}}_{i} || \mathbf{W}\vec{\mathbf{x}}_{j}]\Big)
\end{equation}
where $||$ is the concatenation operator.
Note, the computation for $\beta_{ij}$ implies each word will have an 
impact on every other word in $\mathbf{G}$, which is computationally inefficient and may 
inject noise in the model training. In order to avoid this, we perform masked attention, 
i.e., compute the attention weight $\beta_{ij}$ for a fixed neighborhood of word $i$, 
i.e., $j \in \mathcal{N}_{i}$. We use the softmax function to compute the normalized 
attention coefficients $\alpha_{ij}$, shown as follows:

\begin{equation}
\alpha_{ij} = \text{softmax}(\beta_{ij}) = \frac{\exp(\beta_{ij})}{\sum_{k \in \mathcal{N}_{i}} \exp(\beta_{ik})}
\end{equation}
Finally, we use the normalized coefficients in order to compute a linear combination 
of the corresponding word representations as the final output representation of each word as follows:

\begin{equation}
\vec{\mathbf{x}^{'}_{i}} = \sigma \Big( \sum_{j \in \mathcal{N}_{i}} \alpha_{ij}\mathbf{W}\vec{\mathbf{x}_{i}} \Big)
\end{equation}
where $\sigma$ is a nonlinearity. 

Though~\citet{2017_velivckovic_GAT} extend their work to a multi-head attention setting, 
but for \OurMODEL, we resort to one attention layer in order to avoid the computational overhead.

The intuitive explanation for the graph attention part of \OurMODEL~is to surround
each word by a set of semantically related words by forming edges in the graph and 
re-compute the representation of each word by propagating information from the 
neighbors in a way that it accommodates the impact of semantic variations of each 
word in an attentive way.

\subsubsection{Isomorphism Loss}
Finally, we use the output of the graph attention layer ($\mathbf{X^{'}}$) to compute the 
isomorphism loss for \OurMODEL~relative to the target embeddings $\mathbf{Y}$. For this, 
we analyze the impact of multiple different variants of isomorphism loss functions 
referred to as $\mathcal{L}_{Iso}$. The details about different variants of the isomorphic loss 
functions are as follows:

\paragraph{L2 Loss ($\mathcal{L}_{2}$).}
We use L2-norm averaged over the number of words as our isomorphism metric.
For $N$ words, $\mathcal{L}_{2}$ is computed as:

\begin{equation}
\mathcal{L}_{2} = \frac{1}{N} {||\mathbf{X^{'}}-\mathbf{Y}||_{2}}
\end{equation}

\paragraph{Orthogonal Procrustus Loss ($\mathcal{L}_{proc}$).}
The orthogonal Procrustes problem aims to find a linear transformation $\mathbf{W}_{p}$ to solve the following metric:

\begin{equation}
\label{proc_eq}
\begin{aligned}
\mathcal{L}_{proc} = \argmin_{\mathbf{W}_{p} \in \mathbf{R}^{d \times d}, \mathbf{W}_{p}^{T}\mathbf{W}_{p}=I} \frac{1}{N}{||\mathbf{X^{'}} \mathbf{W}_{p}-\mathbf{Y}||}_{2}
\end{aligned}
\end{equation}

For this, we use an existing solution $\mathbf{W}_{p} = \QQ\PP^{T}$ proposed by~\citet{1966_schonemann}, 
where $\PP \Sigma \QQ^{T}$ is the singular value decomposition 
of the matrix $\YY^{T}\XX^{'}$.

\paragraph{A variant of Procrustus Loss ($\mathcal{L}_{proc_{src}}$).}
For this, we follow the same process flow as outlined above for the Procrustus 
loss. The only difference is that we use pre-trained embeddings for the
target words to initialize the corresponding embeddings for the source words 
for a given set of translation seed pairs.
The end goal of this setting is to analyze the contribution of the pre-trained 
embeddings to guide the overall isomorphism of the source embeddings.
Note that the initialized embeddings for the source words are updated during the 
model training.

\subsection{The Complete Model}
Finally, we combine the loss for the skip-gram distributional 
training objective with the isomorphism loss in order to come 
up with the loss function of \OurMODEL, as shown below:

\begin{equation}
\label{Eq:Final}
\mathcal{L}_{\OurMODEL} = \gamma  \mathcal{L}_{Dis} + (1-\gamma)  \mathcal{L}_{Iso}
\end{equation}

where, $\gamma$ is the hyper-parameter controlling the contribution of 
individual losses in the model.

\vspace{-1ex}
\section{Experiments and Results}
\vspace{-1ex}

\subsection{Datasets}
For comparative analysis, we use the same data settings as primarily used by recent work, 
i.e., IsoVec by~\citet{2022_isovec}. For the main experiments (section~\ref{main_res}), 
we use the first 1 million lines of the newscrawl-2020 data set for the English and 
Arabic languages~\citep{barrault_2020}.
For the domain mismatch settings (section~\ref{dom_res}), we use 33.8 million lines 
of web-crawl data for the English language and newscrawl-2020 data for the Arabic language.
For data pre-processing, we use Moses scripts\footnote{\url{github.com/moses-smt/mosesdecoder/tree/master/scripts/tokenizer}} 
to process the English language data. 
For the Arabic language, we use NLTK tokenizer\footnote{\url{https://www.nltk.org/api/nltk.tokenize.html}}.
For performance evaluation, we used publically available train, dev, and test splits 
provided by MUSE~\citep{muse_2017}. We use word pairs numbered: 0-5000, 5001-6500, 
and 6501-8000 as train, test, and dev splits respectively. The train split is used for model 
training, and dev split for parameter tuning. The final results are computed over the test split.

\subsection{Baseline Models}
We use independently trained distributional embeddings for the source and target 
languages (without the isomorphism loss) as an immediate baseline. 
Other than this, we compare \OurMODEL~against the existing best-performing model 
on relative isomorphism, i.e., IsoVec by~\citet{2022_isovec}.
Note, IsoVec follows a similar approach as that of \OurMODEL~with the 
distinction that~\OurMODEL~uses graph attention as an additional 
layer to control the relative isomorphism of semantically relevant words. 
For IsoVec, we used publicly available implementation provided by the authors 
to generate the results for the Arabic language.

\subsection{Experimental Settings}
In order to train the proposed model, i.e., \OurMODEL, we use Adam 
optimizer~\cite{2014_adam} with learning rate = 0.001. 
In Equation~\ref{SG_eq}, we set the value of $k$ = 10.
In Equation~\ref{Eq:Final}, we use the value of $\gamma$ = 0.333.
For the graph construction process, $\eta = 0.4$.
We use English as the target language, and Arabic as the source language. 
Similar to the baseline models, we use VecMap toolkit (explained in 
Section~\ref{vecmap_bkgd}) for mapping across different embedding spaces. 
We use average precision (i.e., P@1) as our evaluation metric,
and report the mean ($\mu$) and standard deviation ($\sigma$) 
of the results averaged over 5 runs of the experiment. All the experiments 
are performed using Intel Core-i9-10900 CPU and Nvidia 1080Ti GPUs.

\subsection{Main Results}
\label{main_res}
The results of~\OurMODEL~compared against the baseline models are shown in Table~\ref{tab:res1}.
We bold-face overall best scores and underline the previous state-of-the-art.

\begin{table}[t]
	\centering
	\resizebox{0.70\columnwidth}{!}{
		\begin{tabular}{lll|cl}
			\hline
			\multicolumn{3}{l}{Methodology}              & \multicolumn{2}{|c}{Avg. P@1}    \\
			\hline
			\multicolumn{3}{l}{Baseline}                & \multicolumn{2}{|c}{15.58 ($\pm$ 0.8)}                              \\
			\hline
			\multicolumn{3}{l}{IsoVec (L2)}             & \multicolumn{2}{|c}{19.59 ($\pm$ 0.7)}                               \\
			\multicolumn{3}{l}{IsoVec (Proc-L2)}        & \multicolumn{2}{|c}{20.03 ($\pm$ 0.5)}                               \\
			\multicolumn{3}{l}{IsoVec (Proc-L2-Init)}   & \multicolumn{2}{|c}{\underline{22.10 ($\pm$ 0.5)}}                   \\
			\hline
			\multicolumn{3}{l}{\OurMODEL~($\mathcal{L}_2$)}                 & \multicolumn{2}{|c}{29.32 ($\pm$ 0.09)}         \\
			\multicolumn{3}{l}{\OurMODEL~($\mathcal{L}_{{proc}_{src}}$)}   & \multicolumn{2}{|c}{\textbf{31.15 ($\pm$ 0.07)}}         \\
			\multicolumn{3}{l}{\OurMODEL~($\mathcal{L}_{proc}$)}            & \multicolumn{2}{|c}{30.60 ($\pm$ 0.21)} \\
			\hline
	\end{tabular}}
	\caption{The results for the proposed model compared against the baseline model 
		and existing state-of-the-art work on relative isomorphism, i.e., IsoVec~\citep{2022_isovec}.}
	\vspace{-3.7ex}
	\label{tab:res1}
\end{table}

These results show that \OurMODEL~outperforms the baseline models by a significant margin. 
The results of \OurMODEL~with different isomorphism loss functions show that almost all the 
loss functions exhibit a similar performance with the loss ($\mathcal{L}_{proc_{src}}$) 
yielding overall best scores.
Compared with the best performing baseline scores, \OurMODEL($\mathcal{L}_{proc_{src}}$) improves the 
average P@1 by approximately 40.95\%.
For the variants of \OurMODEL~with loss functions $\mathcal{L}_{2}$ and 
$\mathcal{L}_{proc}$ the improvement in performance is 32.67\% and 38.46\% respectively.
A relatively higher performance for the loss $\mathcal{L}_{proc_{src}}$ compared to 
$\mathcal{L}_{proc}$ shows that initializing the source embeddings with corresponding 
translation pairs from the target embeddings had a beneficial impact on the model training.
Analyzing the variance of the results, we observe the variance of \OurMODEL~is much lower compared to the variance of the baseline models. 
The worst-case variance of \OurMODEL~is even less than half of the variance of the 
baseline models, which shows that \OurMODEL~yields an overall stable performance
across multiple re-runs of the experiments.

To summarize, these experiments show the essence of using the graph attention 
layers on controlling the relative isomorphism of the embedding spaces for BLI.
We attribute the performance gained by \OurMODEL~to the ability of the self-attention 
mechanism to appropriately accumulate information from semantically related words, 
which in turn plays a significant role in controlling the relative isomorphism of the embedding spaces.

\vspace{-1ex}
\section{Discussion}
\vspace{-1ex}
In this section, we perform a detailed analysis of~\OurMODEL~under different settings.
For this, we perform analyses encompassing:
(i) domain mismatch settings, 
(ii) correlation with isometric metrics, and
(iii) error analysis.

\subsection{Domain Mis-match}
\label{dom_res}

\begin{table}[t]
	\centering
	\resizebox{0.70\columnwidth}{!}{
		\begin{tabular}{lll|cc}
			\hline
			\multicolumn{3}{l}{Methodology}  & \multicolumn{2}{|c}{Avg. P@1}\\
			\hline
			\multicolumn{3}{l}{Baseline}     & \multicolumn{2}{|c}{14.70 ($\pm$ 0.7)}  \\
			\hline
			\multicolumn{3}{l}{IsoVec (L2)} & \multicolumn{2}{|c}{18.49 ($\pm$ 0.6)} \\
			\multicolumn{3}{l}{IsoVec (Proc-L2)} & \multicolumn{2}{|c}{18.80 ($\pm$ 0.7)} \\
			\multicolumn{3}{l}{IsoVec (Proc-L2-Init)} & \multicolumn{2}{|c}{\underline{19.14 ($\pm$ 0.7)}} \\
			\hline
			\multicolumn{3}{l}{\OurMODEL~($\mathcal{L}_2$)}  & \multicolumn{2}{|c}{29.69 ($\pm$ 0.18)}  \\
			\multicolumn{3}{l}{\OurMODEL~($\mathcal{L}_{{proc}_{src}}$)}   & \multicolumn{2}{|c}{32.27 ($\pm$ 0.17)}  \\
			\multicolumn{3}{l}{\OurMODEL~($\mathcal{L}_{proc}$)} & \multicolumn{2}{|c}{\textbf{33.84 ($\pm$ 0.02)}}  \\
			\hline
	\end{tabular}}
	\caption{The results for the proposed approach under domain mismatch settings compared against the 
		baseline model and existing state-of-the-art work on relative isomorphism, i.e., IsoVec~\citep{2022_isovec}.}
	\label{tab:domain}
\end{table}

The results of our model for domain mismatch settings are shown in Table~\ref{tab:domain}. 
Similar to the results for the main experiments, we also compare these 
results against the baseline models. We boldface the overall best scores with existing state-of-the-art 
underlined. These results show that \OurMODEL~yields higher performance compared to the baseline 
models. The variants of \OurMODEL~with loss $\mathcal{L}_2$, $\mathcal{L}_{proc}$ and 
$\mathcal{L}_{proc_{src}}$ outperform the best performing baseline model by 55.12\%, 
76.80\%, and 68.60\% respectively.

Comparing these results to the results for the main experiments (reported in Table~\ref{tab:res1}), 
we observe that \OurMODEL~yields a better performance for the domain mismatch settings 
relative to the in-domain setting.
We attribute this performance improvement to: 
(a) the ability of \OurMODEL~to capture and consolidate information from semantically 
relevant words even from different domains, 
(b) a relatively larger corpus for the target language (English) for domain mismatch settings.
We notice that in contrast to the main experiments, for the domain mismatch settings loss 
the model \OurMODEL($\mathcal{L}_{proc}$) yields a better performance compared to 
\OurMODEL($\mathcal{L}_{proc_{src}}$). This shows that with the increase in the size of 
the data, the capability of the graph attention part of \OurMODEL~to accumulate information 
about the semantically related words augments in a way that it even surpasses the model 
training with seed embeddings initialized.

Note, as illustrated in Section~\ref{intro}, domain mismatch is a key challenge for the BLI 
systems. Earlier research by~\citet{2018_sogaard} shows that the majority of existing BLI systems 
perform poorly in inferring bilingual information from embeddings trained on different data 
domains.
One key challenge that hinders the performance of these BLI systems is their inability to 
incorporate the impact of semantically related keywords and/or jargons peculiarly related 
to different domains.
These words though belonging to different data domains have similar meanings and 
BLI systems should appropriately use this information for the model training. 
This makes \OurMODEL~a better alternate, especially because of its provision to accumulate information about 
multiple different semantically related words using graph attention layers, as is also 
evident by a relatively higher performance of \OurMODEL~compared to the baseline models.


\begin{table}[t]
	\centering
	\resizebox{0.70\columnwidth}{!}{
		\begin{tabular}{l|c|c}
			\hline
			& \multicolumn{1}{c}{ ES ($\downarrow$)} & \multicolumn{1}{|c}{$\rho$ ($\uparrow$)} \\
			\hline
			\OurMODEL~($\mathcal{L}_2$)             & 80.99     & 0.46        \\
			\OurMODEL~($\mathcal{L}_{proc}$)        & 99.89     & \textbf{0.56}        \\
			\OurMODEL~($\mathcal{L}_{proc_{src}}$)  & \textbf{76.89}     & 0.45         \\
			\hline
	\end{tabular}}
	\caption{Analysis of different isometry metrics for~\OurMODEL, i.e., , Eigenvector Similarity (ES) 
		and Pearson's Correlation ($\rho$).}
	\vspace{-3.7ex}
	\label{tab:isometric}
\end{table}

\subsection{Correlation with isometric metrics}
Similar to the existing works on controlling the relative isomorphism of the embedding spaces~\citep{2022_isovec}, 
we compute isomorphism metrics for the results of \OurMODEL.
We use two widely used metrics, namely: (i) Eigenvector similarity, (ii) Pearson's correlation.
The computation details, and results of \OurMODEL~for these metrics are as follows:

\paragraph{Eigenvector Similarity (ES).}
In order to compute the eigenvector similarity between the embedding spaces, 
we compute the Laplacian spectra of corresponding k-nearest neighbour graphs.
We expect the graphs with similar structures to have similar eigenvalue spectra.
For this, we follow the same settings as that of~\citet{2018_sogaard}. Given the seed 
pairs $\{x_0, x_1,..., x_s\}$ and $\{y_0, y_1,..., y_s\}$, we proceed as follows:
(i) compute unweighted k-nearest neighbour graphs (i.e., $\text{G}_{X}$ and $\text{G}_{Y}$),
(ii) compute the graph Laplacians $L_{\text{G}_{X}}$ and $L_{\text{G}_{Y}}$, where $L_{\text{G}} = D_{\text{G}} - A_{\text{G}}$,
(iii) compute the eigenvalues for each graph Laplacian, i.e., \{$\lambda_{L_{\text{G}_{X}}}(i)$; $\lambda_{L_{\text{G}_{Y}}}(i)$\}
(iv) select $r = min(r_{X}, r_{Y})$ where $r_X$ is the maximum $r$ such that the first $r$ 
eigenvalues of $L_{\text{G}_{X}}$ sum to less than 90\% of the total sum of the eigenvalues.
(v) depending upon the value of $r$, compute the eigenvector similarity as:
$\sum_{i=1}^{r} (\lambda_{L_{\text{G}_{X}}}(i) - \lambda_{L_{\text{G}_{Y}}}(i))^{2}$.

The results for the eigenvector similarity measures should have an inverse correlation ($\downarrow$)
with the P@1. The results in the left column of Table~\ref{tab:isometric} show that
the variant of \OurMODEL~with loss $\mathcal{L}_{proc_{src}}$ yields a higher performance
which aligns with our findings for the main experiments in Table~\ref{tab:res1}.
However, the ES scores for the model with $\mathcal{L}_{2}$ and $\mathcal{L}_{proc}$
show irregular behavior. 
We expect the model with the loss $\mathcal{L}_{proc}$ to have
a lower value for the ES score compared to $\mathcal{L}_{2}$, which is in 
contrast to our findings in Table~\ref{tab:isometric}.

\paragraph{Pearson's Correlation($\rho$).}
In order to calculate the Pearson's correlation, we first compute the pairwise cosine
similarity scores for the seed translation pairs, i.e., 
$\{\cos(x_0,x_1), \cos(x_0,x_2),..., \cos(x_s,x_s)\}$, and 
$\{\cos(y_0,y_1), \cos(y_0,y_2),..., \cos(y_s,y_s)\}$. Later, we compute 
the Pearson’s correlation between the lists of cosine similarity scores.
We expect the Pearson's correlation score to correlate positively ($\uparrow$) 
with the average P@1.

The results in the right half of Table~\ref{tab:isometric} show the
Pearson's correlation scores for all the variants of \OurMODEL.
These results show an unclear behaviour, with $\mathcal{L}_{proc}$ showing
better performance compared to $\mathcal{L}_2$ and $\mathcal{L}_{proc_{src}}$.
This is in contrast to the results for P@1 reported in Table~\ref{tab:res1}, where 
$\mathcal{L}_{proc_{src}}$ shows a better performance compared to other models.

To summarize our findings for the isometric metrics, we observe that these 
results do not truly correlate with the average P@1.
These findings are consistent with the earlier study IsoVec~\cite{2022_isovec} 
that also emphasized the need for better isomorphism metrics in order to portray 
the correct picture of the degree of relative isomorphism of the 
embedding spaces.

\subsection{Error Analysis}
In this section, we perform a detailed analysis of the error cases of \OurMODEL~in 
order to know:
(i) the performance improvement attributable to the graph attention part of the model, 
(ii) limitations of the \OurMODEL,~and room for potential improvement.
For this, we perform error analysis on two variants of \OurMODEL, i.e., with and 
without graph attention layer. 
All experiments are performed using the in-domain settings using the 
best-performing model, i.e., \OurMODEL~($\mathcal{L}_{proc_{src}}$). 
Details are as follows:

\paragraph{\OurMODEL~(w/o Graph Attention).}
We initially analyze the error cases for the basic variant of 
\OurMODEL~(without the graph attention layer) that have been 
corrected by the complete model.
The core focus of this analysis is to look for the translation 
instances that benefit especially due to the graph attention mechanism.
Note, for this analysis, we only include error cases that have incorrect prediction 
for the basic model (i.e., without graph attention) and are correctly classified by the 
complete model~\OurMODEL.

\begin{table}[t]
	\centering
	\resizebox{0.70\columnwidth}{!}{
		\begin{tabular}{c|c|c}
			\hline
			\multicolumn{3}{c}{\OurMODEL~(w/o Graph Attention)} \\
			\hline
			\multicolumn{1}{c}{source} & \multicolumn{1}{c}{target$^{'}$} & \multicolumn{1}{c}{target} \\
			\hline
			\<الليزر>  & infrared     & laser   \\
			\<الفهم>  & pronunciation & understanding   \\
			\<السجل>  & database      & register   \\
			\<تلفاز>  & keyboards     & tv   \\
			\<صدى>    & elated        & echo   \\
			\<اربعة>  & three         & four   \\
			\<زرقاء>  & foreboding    & blue   \\
			\hline
	\end{tabular}}
	\caption{Example error cases for the model: \OurMODEL~(w/o Graph Attention). The ``target$^{'}$'' 
		represents the model predictions, ``target'' represents the ground truth.}
	\vspace{-2.7ex}
	\label{tab:error1}
\end{table}

While the graph attention layer is able to correct approximately 11\% of the errors made 
by the basic variant of \OurMODEL, we observe almost 72\% of the error cases belong 
to the noun category. 
One possible explanation in this regard is that the phenomenon of multiple senses is more 
dominant among the nouns in contrast to other parts-of-the speech, e.g., verbs and adjectives,
which makes it harder to control their relative isomorphism~\cite{2019_ANTSYN}. 
Some examples in this regard have been shown in Table~\ref{tab:error1}.
We also observe that the majority of the predictions made by the basic variant of 
\OurMODEL~are not semantically related to the true target words, which clearly indicates the need 
for information sharing among the semantically related words required to control 
the relative isomorphism of the embedding spaces.

\paragraph{\OurMODEL~(The Complete Model).}
The end goal of performing error analysis on the complete model is to dig out the 
potential reasons and/or understanding of the limitations of the proposed model.
Note, we perform this analysis for the best-performing variant of \OurMODEL, i.e., 
with the loss $\mathcal{L}_{proc_{src}}$.

We randomly select a subset of 50 error cases for quantification.
To our surprise, most of the errors (approximately 65\%) made by \OurMODEL~are either 
semantically very close to the true target word or a lexical variant of the true target word.
Some examples in this regard have been shown in Table~\ref{tab:error2}.
These results clearly show the current performance of \OurMODEL~is underrated primarily 
due to the use of a very strict evaluation criterion, (i.e., P@1).
This calls for the need for better and more sophisticated mechanisms for the BLI 
systems in order to measure the relative isomorphism of the geometric spaces.

\begin{table}[t]
	\centering
	\resizebox{0.65\columnwidth}{!}{
		\begin{tabular}{c|c|c}
			\hline
			\multicolumn{3}{c}{\OurMODEL~($\mathcal{L}_{proc_{src}}$)} \\
			\hline
			\multicolumn{1}{c}{source} & \multicolumn{1}{c}{target$^{'}$} & \multicolumn{1}{c}{target} \\
			\hline
			\<ضرورية>     & vital             & necessary  \\
			\<شمل>        & includes          & included  \\
			\<القلم>      & pencil            & pen  \\
			\<سماع>       & hear              & hearing  \\
			\<المواهب>    & talents           & talent  \\
			\<الفولاذ>     & metal             & steel  \\
			\<المواصفات>  & certifications    & specs  \\
			\hline
	\end{tabular}}
	\caption{Example error cases for \OurMODEL~using the loss function $\mathcal{L}_{proc_{src}}$. 
		The ``target$^{'}$'' represents the model predictions, and the ``target'' represents the ground truth.}
	\vspace{-2.7ex}
	\label{tab:error2}
\end{table}

To summarize, the error analysis shows the essence of using graph attention in 
order to control the relative isomorphism of the embedding spaces. It helps in 
incorporating and/or accumulating information across semantically related words 
in order to perform the end task in a robust way. 

\vspace{-1ex}
\section{Conclusions and Future Research}
\vspace{-1ex}
In this work, we propose Graph Attention for Relative Isomorphism (\OurMODEL).
\OurMODEL~incorporates the impact of semantically related words in order to control the relative isomorphism of geometric spaces in a performance-enhanced way. Experimental evaluation using the Arabic data set shows that 
\OurMODEL~outperforms the existing state-of-the-art research
by 40.95\% and 76.80\% for in-domain and domain mismatch settings.
In the future, we will extend this research to deep contextualized 
embeddings and non-euclidean geometries.

\vspace{-1ex}
\section{Limitations}
\vspace{-1ex}
Some of the core limitations of the proposed approach are outlined as follows:
(i) all the techniques have been developed assuming a Euclidean 
geometry for the underlying embedding spaces, its extension to non-Euclidean 
spaces are still unaddressed, (ii) the existing problem formulation is not 
defined for the deep contextualized embeddings.

\paragraph*{Acknowledgements.}%
Di Wang, Yan Hu and Muhammad Asif Ali are supported in part by the baseline funding BAS/1/1689-01-01, funding from the CRG grand URF/1/4663-01-01, FCC/1/1976-49-01 from CBRC and funding from the AI Initiative REI/1/4811-10-01 of King Abdullah University of Science and Technology (KAUST). Di Wang is also supported by the funding of the SDAIA-KAUST Center of Excellence in Data Science and Artificial Intelligence (SDAIA-KAUST AI).

\bibliography{anthology,custom}
\bibliographystyle{acl_natbib}


\end{document}